# Anytime State-Based Solution Methods for Decision Processes with non-Markovian Rewards

**Sylvie Thiébaux**
Computer Sciences Laboratory
The Australian National University
Canberra, ACT, Australia

**Froduald Kabanza**
Dept. Mathématiques et Informatique
Université de Sherbrooke
Sherbrooke, Québec, Canada

**John Slaney**
Computer Sciences Laboratory
The Australian National University
Canberra, ACT, Australia

## Abstract

A popular approach to solving a decision process with non-Markovian rewards (NMRDP) is to exploit a compact representation of the reward function to automatically translate the NMRDP into an equivalent Markov decision process (MDP) amenable to our favorite MDP solution method. The contribution of this paper is a representation of non-Markovian reward functions and a translation into MDP aimed at making the best possible use of state-based anytime algorithms as the solution method. By explicitly constructing and exploring only parts of the state space, these algorithms are able to trade computation time for policy quality, and have proven quite effective in dealing with large MDPs. Our representation extends future linear temporal logic (FLTL) to express rewards. Our translation has the effect of embedding model-checking in the solution method. It results in an MDP of the minimal size achievable without stepping outside the anytime framework, and consequently in better policies by the deadline.

## 1 INTRODUCTION

Markov decision processes (MDPs) are now widely accepted as the preferred model for decision-theoretic planning problems (Boutilier et al., 1999). The fundamental assumption behind the MDP formulation is that not only the system dynamics but also the reward function are *Markovian*. Therefore, all information needed to determine the reward at a given state must be encoded in the state itself.

This requirement is not always easy to meet for planning problems, as many desirable behaviors are naturally expressed as properties of execution *sequences*, see e.g. (Drummond, 1989; Haddawy and Hanks, 1992; Bacchus and Kabanza, 1998; Pistore and Traverso, 2001). Typical cases include rewards for the maintenance of some property, for the periodic achievement of some goal, for the achievement of a goal within a given number of steps of the request being made, or even simply for the very first achievement of a goal which becomes irrelevant afterwards. A decision process in which rewards depend on the sequence of states passed through rather than merely on the current state is called a decision process with *non-Markovian rewards* (NMRDP) (Bacchus et al., 1996).

A difficulty with NMRDPs is that the most efficient MDP solution methods do not directly apply to them. The traditional way to circumvent this problem is to formulate the NMRDP as an equivalent MDP, whose states are those of the underlying system expanded to encode enough history-dependent information to determine the rewards. Hand crafting such an *expanded* MDP (XMDP) can however be very difficult in general. This is exacerbated by the fact that the size of the XMDP impacts on the effectiveness of many solution methods. Therefore, there has been interest in automating the translation into an XMDP, starting from a natural specification of non-Markovian rewards and of the system's dynamics (Bacchus et al., 1996; Bacchus et al., 1997). This is the problem we focus on.

When solving NMRDPs in this setting, the central issue is to define a non-Markovian reward specification language and a translation into an XMDP *adapted* to the class of MDP solution methods and representations we would like to use for the type of problems at hand. The two previous proposals within this line of research both rely on past linear temporal logic (PLTL) formulae to specify the behaviors to be rewarded (Bacchus et al., 1996; Bacchus et al., 1997), but adopt two very different translations adapted to two very different types of solution methods and representations. The translation in (Bacchus et al., 1996) targets classical *state-based* solution methods such as policy iteration (Howard, 1960) which generate *complete* policies at the cost of enumerating all states in the entire MDP, while that in (Bacchus et al., 1997) targets *structured* solution methods and representations, which do not require explicit state enumeration, see e.g. (Boutilier et al., 2000).

The aim of the present paper is to provide a language and a translation adapted to another class of solution methods which have proven quite effective in dealing with large MDPs, namely *anytime* state-based methods such as (Barto et al., 1995; Dean et al., 1995; Thiébaux et al., 1995; Hansen and Zilberstein, 2001). These methods typically start with a compact representation of the MDP based on probabilistic planning operators, and search forward from an initial state, constructing new states by expanding the envelope of the policy as time permits. They may produce an approximate and even incomplete policy, but only explicitly construct and explore a fraction of the MDP. Neither of the two previous proposals is well-suited to such solution methods, the first because the cost of the translation (most of which is performed prior to the solution phase) annihilates the benefits of anytime algorithms, and the second because the size of the XMDP obtained is an obstacle to the applicability of state-based methods. Since here both the cost of the translation and the size of the XMDP it results in will severely impact on the quality of the policy obtainable by the deadline, we need an appropriate resolution of the tradeoff between the two.

Our approach has the following main features. The translation is entirely embedded in the anytime solution method, to which full control is given as to which parts of the XMDP will be explicitly constructed and explored. While the XMDP obtained is not minimal, it is of the minimal size achievable without stepping outside of the anytime framework, i.e., without enumerating parts of the state or expanded state spaces that the solution method would not necessarily explore. This relaxed notion of minimality, which we call *blind minimality* is the most appropriate in the context of anytime state-based solution methods.

When the rewarding behaviors are specified in PLTL, there does not appear to be a way of achieving a relaxed notion of minimality as powerful as blind minimality without a prohibitive translation. Therefore instead of PLTL, we adopt a variant of *future* linear temporal logic (FLTL) as our specification language, which we extend to handle rewards. While the language has a more complex semantics than PLTL, it enables a natural translation into a blind-minimal XMDP by simple *progression* of the reward formulae. Moreover, search control knowledge expressed in FLTL (Bacchus and Kabanza, 2000) fits particularly nicely in this model-checking framework, and can be used to dramatically reduce the fraction of the search space explored by anytime solution methods.

The paper is organised as follows. Section 2 begins with background material on MDPs, NMRDPs, XMDPs, and anytime state-based solution methods. Section 3 describes our language for specifying non-Markovian rewards and the progression algorithm. Section 4 defines our translation into an XMDP along with the concept of blind minimality it achieves, and presents our approach to the embedded construction and solution of the XMDP. Finally, Section 5, provides a detailed comparison with previous approaches, and concludes with some remarks about future work. The proofs of the theorems appear in (Thiébaux et al., 2002).

## 2 BACKGROUND

### 2.1 MDPs

A Markov decision process of the type we consider is a 5-tuple $\langle S, s_0, A, \Pr, R \rangle$, where $S$ is a finite set of fully observable states, $s_0 \in S$ is the initial state, $A$ is a finite set of actions ($A(s)$ denotes the subset of actions applicable in $s \in S$), $\{\Pr(s, a, \bullet) \mid s \in S, a \in A(s)\}$ is a family of probability distributions over $S$, such that $\Pr(s, a, s')$ is the probability of being in state $s'$ after performing action $a$ in state $s$, and $R : S \mapsto \mathbb{R}$ is a reward function such that $R(s)$ is the immediate reward for being in state $s$. It is well known that such an MDP can be compactly represented using probabilistic extensions of traditional planning languages, see e.g., (Kushmerick et al., 1995; Thiébaux et al., 1995).

A stationary policy for an MDP is a function $\pi : S \mapsto A$, such that $\pi(s) \in A(s)$ is the action to be executed in state $S$. We note $E(\pi)$ the envelope of the policy, that is the set of states that are reachable (with a non-zero probability) from the initial state $s_0$ under the policy. If $\pi$ is defined at all $s \in E(\pi)$, we say that the policy is complete, and that it is incomplete otherwise. We note $F(\pi)$ the set of states in $E(\pi)$ at which $\pi$ is undefined. $F(\pi)$ is called the fringe of the policy. We stipulate that the fringe states are absorbing. The value of a policy $\pi$ at any state $s \in E(\pi)$, noted $V_\pi(s)$ is the sum of the expected rewards to be received at each future time step, discounted by how far into the future they occur. That is, for a non-fringe state $s \in E(\pi) \setminus F(\pi)$:

$$V_\pi(s) = R(s) + \beta \sum_{s' \in E(\pi)} \Pr(s, \pi(s), s') V_\pi(s')$$

where $0 \leq \beta \leq 1$ is the discounting factor controlling the contribution of distant rewards. For a fringe state $s \in F(\pi)$, $V_\pi(s)$ is heuristic or is the value at $s$ of a complete default policy to be executed in absence of an explicit one. For the type of MDP we consider, the value of a policy $\pi$ is the value $V_\pi(s_0)$ of $\pi$ at the initial state $s_0$, and the larger this value, the better the policy.

### 2.2 STATE-BASED ANYTIME ALGORITHMS

Traditional state-based solution methods such as policy iteration (Howard, 1960) can be used to produce an optimal complete policy. Policy iteration can also be viewed as an anytime algorithm, which returns a complete policy whose value increases with computation time and converges to optimal. The main drawback of policy iteration is that it explicitly enumerates all states that are reachable from $s_0$ in the entire MDP. Therefore, there has been interest in other anytime solution methods, which may produce incomplete policies, but only enumerate an increasing fraction of the states policy iteration requires.

For instance, (Dean et al., 1995) describes methods which deploy policy iteration on judiciously chosen larger and

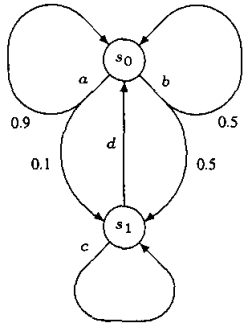

In the initial state $s_0$, $p$ is false and two actions are possible: $a$ causes a transition to $s_1$ with probability 0.1, and no change with probability 0.9, while for $b$ the transition probabilities are 0.5. In state $s_1$, $p$ is true, and actions $c$ and $d$ ("stay" and "go") lead to $s_1$ and $s_0$ with probability 1.
A reward is received the first time $p$ is true, but not subsequently. That is, the rewarded state sequences are:
$\langle s_0, s_1 \rangle$
$\langle s_0, s_0, s_1 \rangle$
$\langle s_0, s_0, s_0, s_1 \rangle$
etc.

Figure 1: A simple NMRDP

larger envelopes. Another example is (Thiébaux et al., 1995), in which a backtracking forward search in the space of (possibly incomplete) policies rooted at $s_0$ is performed until interrupted, at which point the best policy found so far is returned. Real-time dynamic programming (RTDP) (Barto et al., 1995), is another popular anytime algorithm, which is to MDPs what learning real-time A* (Korf, 1990) is to deterministic domains. It can be run on-line, or off-line for a given number of steps or until interrupted. A more recent example is the LAO* algorithm (Hansen and Zilberstein, 2001) which combines dynamic programming with heuristic search.

All these algorithms eventually converge to the optimal policy but need not necessarily explore the entire state space to guarantee optimality.[1] When interrupted before convergence, they return a possibly incomplete but often useful policy. Another common point of these approaches is that they perform a forward search, starting from $s_0$ and repeatedly expanding the envelope of the current policy one step forward. Since planning operators are used to compactly represent the state space, these methods will only explicitly construct a subset of the MDP. In this paper, we will use these solution methods to solve decision processes with non-Markovian rewards which we define next.

### 2.3 NMRDPs AND EQUIVALENT XMDPs

We first need some notation. Let $S^*$ be the set of finite sequences of states over $S$, and $S^\omega$ be the set of possibly infinite state sequences. In the following, where '$\Gamma$' stands for a possibly infinite state sequence in $S^\omega$ and $i$ is a natural number, by '$\Gamma_i$' we mean the state of index $i$ in $\Gamma$, by '$\Gamma(i)$' we mean the prefix $\langle \Gamma_0, \ldots, \Gamma_i \rangle \in S^*$ of $\Gamma$, and by $\text{pre}(\Gamma)$ we mean the set of finite prefixes of $\Gamma$. $\Gamma_1; \Gamma_2$ denotes the concatenation of $\Gamma_1 \in S^*$ and $\Gamma_2 \in S^\omega$. For a decision process $D = \langle S, s_0, A, \Pr, R \rangle$ and a state $s \in S$, $\widetilde{D}(s)$ stands for the set of state sequences rooted at $s$ that are feasible under the actions in $D$, that is: $\widetilde{D}(s) = \{\Gamma \in S^\omega \mid \Gamma_0 = s \text{ and } \forall i\ \exists a \in A(\Gamma_i)\ \Pr(\Gamma_i, a, \Gamma_{i+1}) > 0\}$. Note that the definition of $\widetilde{D}(s)$ does not depend on $R$ and therefore also stands for NMRDPs which we describe now.

A decision process with non-Markovian rewards is identical to an MDP except that the domain of the reward function is $S^*$. The idea is that if the process has passed through state sequence $\Gamma(i)$ up to stage $i$, then the reward $R(\Gamma(i))$ is received at stage $i$. Figure 1 gives an example. Like the reward function, a policy for an NMRDP depends on history, and is a mapping from $S^*$ to $A$. As before, the value of policy $\pi$ is the expectation of the discounted cumulative reward over an infinite horizon:

$$V_\pi(s_0) = \lim_{n \to \infty} \mathsf{E}\left[\sum_{i=0}^n \beta^i R(\Gamma(i)) \mid \pi, \Gamma_0 = s_0\right]$$

The clever algorithms developed to solve MDPs cannot be directly applied to NMRDPs. One way of dealing with this problem is to formulate the NMRDP as an equivalent MDP with an expanded state space (Bacchus et al., 1996). The expanded states in this XMDP (e-states, for short) augment the states of the NMRDP by encoding additional information sufficient to make the reward history-independent. An e-state can be seen as labeled by a state of the NMRDP (via the function $\tau$ in Definition 1 below) and by history information. The dynamics of NMRDPs being Markovian, the actions and their probabilistic effects in the XMDP are exactly those of the NMRDP. The following definition, adapted from (Bacchus et al., 1996), makes this concept of equivalent XMDP precise. Figure 2 gives an example.

**Definition 1** *MDP $D' = \langle S', s'_0, A', \Pr', R' \rangle$ is an equivalent expansion (or XMDP) for NMRDP $D = \langle S, s_0, A, \Pr, R \rangle$ if there exists a mapping $\tau : S' \mapsto S$ such that:*

1. *$\tau(s'_0) = s_0$.*

2. *For all $s' \in S'$, $A'(s') = A(\tau(s'))$.*

3. *For all $s_1, s_2 \in S$, if there is $a \in A(s_1)$ such that $\Pr(s_1, a, s_2) > 0$, then for all $s'_1 \in S'$ such that $\tau(s'_1) = s_1$, there exists a unique $s'_2 \in S'$, $\tau(s'_2) = s_2$, such that for all $a \in A'(s'_1)$, $\Pr'(s'_1, a, s'_2) = \Pr(s_1, a, s_2)$.*

4. *For any feasible state sequence $\Gamma \in \widetilde{D}(s_0)$ and $\Gamma' \in \widetilde{D}'(s'_0)$ such that $\tau(\Gamma'_i) = \Gamma_i$ for all $i$, we have: $R'(\Gamma'_i) = R(\Gamma(i))$ for all $i$.*

Items 1–3 ensure that there is a bijection between feasible state sequences in the NMRDP and feasible e-state sequences in the XMDP. Therefore, a stationary policy for the XMDP can be reinterpreted as a non-stationary policy for the NMRDP. Furthermore, item 4 ensures that the two policies have identical values, and that consequently, solving an NMRDP optimally reduces to producing an equivalent XMDP and solving it optimally (Bacchus et al., 1996):

**Proposition 1** *Let $D$ be an NMRDP, $D'$ an equivalent XMDP for it, and $\pi'$ a policy for $D'$. Let $\pi$ be the function defined on the sequence prefixes $\Gamma(i) \in \widetilde{D}(s_0)$ by $\pi(\Gamma(i)) = \pi'(\Gamma'_i)$, where for all $j \leq i\ \tau(\Gamma'_j) = \Gamma_j$. Then $\pi$ is a policy for $D$ such that $V_\pi(s_0) = V_{\pi'}(s'_0)$.*

---
[1] This is also true of the basic envelope expansion algorithm in (Dean et al., 1995), under the same conditions as for LAO*.

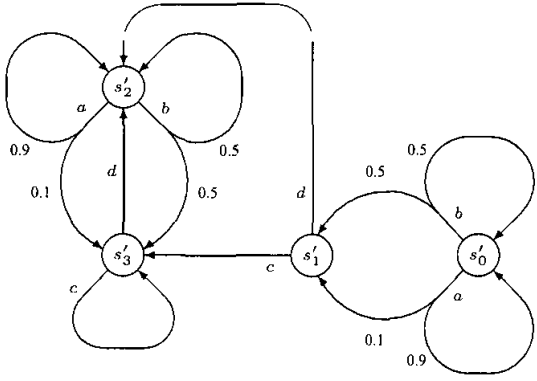

Figure 2: An XMDP equivalent to the NMRDP in Figure 1. $\tau(s'_0) = \tau(s'_2) = s_0$. $\tau(s'_1) = \tau(s'_3) = s_1$. State $s'_1$ is rewarded; the other three states are not.

When solving NMRDPs in this setting, the two key issues are how to specify non-Markovian reward functions compactly, and how to exploit this compact representation to automatically translate an NMRDP into an equivalent XMDP amenable to our favorite solution methods. The goal of this paper is to provide a reward function specification language and a translation that are adapted to the anytime state-based solution methods previously mentioned. We take these problems in turn in the next two sections.

## 3 REWARDING BEHAVIORS

### 3.1 LANGUAGE AND SEMANTICS

Representing non-Markovian reward functions compactly reduces to compactly representing the behaviors of interest, where by *behavior* we mean a set of finite sequences of states (a subset of $S^*$), e.g. the $\{\langle s_0, s_1 \rangle, \langle s_0, s_0, s_1 \rangle, \langle s_0, s_0, s_0, s_1 \rangle \ldots\}$ in Figure 1. Recall that we get rewarded at the end of any prefix $\Gamma(i)$ in that set. Once behaviors are compactly represented, it is straightforward to represent non-Markovian reward functions as mappings from behaviors to real numbers – we shall defer looking at this until Section 3.5.

To represent behaviors compactly, we adopt a version of future linear temporal logic (FLTL) augmented with a propositional constant '\$', intended to be read 'The behavior we want to reward has just happened' or 'The reward is received now'. The language \$FLTL begins with a set of basic propositions $\mathcal{P}$ giving rise to literals:

$$\mathcal{L} ::= \mathcal{P} \mid \neg \mathcal{P} \mid \top \mid \bot \mid \$$$

where $\top$ and $\bot$ stand for 'true' and 'false', respectively. The connectives are classical $\wedge$ and $\vee$, and the temporal modalities $\bigcirc$ (next) and $\mathsf{U}$ (*weak* until), giving formulae:

$$\mathcal{F} ::= \mathcal{L} \mid \mathcal{F} \wedge \mathcal{F} \mid \mathcal{F} \vee \mathcal{F} \mid \bigcirc \mathcal{F} \mid \mathcal{F} \mathsf{U} \mathcal{F}$$

Because our 'until' is weak ($f_1 \mathsf{U} f_2$ means $f_1$ will be true from now on until $f_2$ is, if ever), we can define the useful operator $\Box$ (always): $\Box f \equiv f \mathsf{U} \bot$ ($f$ will always be true from now on). We also adopt the notations $\bigcirc^k f$ for $k$ iterations of the $\bigcirc$ modality ($f$ will be true in exactly $k$ steps), $\bigcirc^{\leq k} f$ for the disjunction of $\bigcirc^i$ for $1 \leq i \leq k$ ($f$ will be true within the next $k$ steps), and $\bigcirc_{\leq k} f$ for the conjunction of $\bigcirc^i f$ for $1 \leq i \leq k$ ($f$ will be true at all the next $k$ steps).

Although negation officially occurs only in literals, i.e., the formulae are in negation formal form (NNF), we allow ourselves to write formulae involving it in the usual way, provided that they have an equivalent in NNF. Not every formula has such an equivalent, because there is no such literal as $\neg \$$ and because eventualities ('f will be true some time') are not expressible. These restrictions are deliberate.

The semantics of this language is similar to that of FLTL, with an important difference: unlike the interpretation of the propositional constants in $\mathcal{P}$, which is fixed (i.e. each state is a fixed subset of $\mathcal{P}$), the interpretation of the constant \$ is not. Remember that \$ means 'The behavior we want to reward has just happened'. Therefore the interpretation of \$ depends on the behavior $B$ we want to reward (whatever that is), and consequently the modelling relation $\models$ stating whether a formula holds at the $i$-th stage of an arbitrary sequence $\Gamma \in S^\omega$, is indexed by $B$. Defining $\models_B$ is the first step in our description of the semantics:

- $(\Gamma, i) \models_B \$ $ iff $\Gamma(i) \in B$
- $(\Gamma, i) \models_B \top$
- $(\Gamma, i) \not\models_B \bot$
- $(\Gamma, i) \models_B p$, for $p \in \mathcal{P}$, iff $p \in \Gamma_i$
- $(\Gamma, i) \models_B \neg p$, for $p \in \mathcal{P}$, iff $p \notin \Gamma_i$
- $(\Gamma, i) \models_B f_1 \wedge f_2$ iff $(\Gamma, i) \models_B f_1$ and $(\Gamma, i) \models_B f_2$
- $(\Gamma, i) \models_B f_1 \vee f_2$ iff $(\Gamma, i) \models_B f_1$ or $(\Gamma, i) \models_B f_2$
- $(\Gamma, i) \models_B \bigcirc f$ iff $(\Gamma, i+1) \models_B f$
- $(\Gamma, i) \models_B f_1 \mathsf{U} f_2$ iff $\forall k \geq i$
  if $(\forall j, i \leq j \leq k \ (\Gamma, j) \not\models_B f_2)$ then $(\Gamma, k) \models_B f_1$

Note that except for subscript $B$ and for the first rule, this is just the standard FLTL semantics, and that therefore \$-free formulae keep their FLTL meaning. As with FLTL, we say $\Gamma \models_B f$ iff $(\Gamma, 0) \models_B f$, and $\models_B f$ iff $\Gamma \models_B f$ for all $\Gamma \in S^\omega$.

The modelling relation $\models_B$ can be seen as specifying when a formula holds, on which reading it takes $B$ as input. Our next and final step is to use the $\models_B$ relation to define, for a formula $f$, the behavior $B_f$ that it represents, and for this we must rather *assume* that $f$ holds, and then *solve* for $B$. For instance, let $f$ be $\Box(p \rightarrow \$)$, i.e., we get rewarded every time $p$ is true. We would like $B_f$ to be the set of all finite sequences ending with a state containing $p$. For an arbitrary $f$, we take $B_f$ to be the set of prefixes that *have* to be rewarded if $f$ is to hold in all sequences:

**Definition 2** $B_f \equiv \bigcap \{B \mid \models_B f\}$



To understand Definition 2, recall that $B$ contains prefixes at the end of which we get a reward and $ evaluates to true. Since $f$ is supposed to describe the way rewards will be received in an *arbitrary* sequence, we are interested in behaviors $B$ which make $ true in such a way as to make $f$ hold regardless of the sequence considered. However, there may be many behaviors with this property, so we take their intersection,[2] ensuring that $B_f$ will only reward a prefix if it has to because that prefix is in *every* behavior satisfying $f$. In all but pathological cases (see Section 3.3), this makes $B_f$ coincide with the (set-inclusion) minimal behavior $B$ such that $\models_B f$. The reason for this 'stingy' semantics, making rewards minimal, is that $f$ does not actually say that rewards are allocated to more prefixes than are required for its truth. For instance, $\Box(p \rightarrow \$)$ *says* only that a reward is given every time $p$ is true, even though a more generous distribution of rewards would be *consistent* with it.

### 3.2 EXAMPLES

It is intuitively clear that many behaviors can be specified by means of $FLTL formulae. There is a list in (Bacchus et al., 1996) of behaviors expressible in PLTL which it might be useful to reward. All of those examples are expressible naturally in $FLTL, as follows.

A simple example is the classical goal formula $g$ saying that a goal $p$ is rewarded whenever it happens: $\Box(p \rightarrow \$)$. As mentioned earlier, $B_g$ is the set of finite sequences of states such that $p$ holds in the last state. If we only care that $p$ is achieved once and get rewarded at each state from then on, we write $\Box(p \rightarrow \Box\$)$. The behavior that this formula represents is the set of finite state sequences having at least one state in which $p$ holds. By contrast, the formula $\neg p \, \mathsf{U} \, (p \wedge \$)$ stipulates that only the first occurrence of $p$ is rewarded (i.e. it specifies the behavior in Figure 1). To reward the occurrence of $p$ at most once every $k$ steps, we write $\Box((\bigcirc^{k+1} p \wedge \neg \bigcirc^{\leq k} p) \rightarrow \bigcirc^{k+1} \$)$.

For response formulae, where the achievement of $p$ is triggered by the command $c$, we write $\Box(c \rightarrow \bigcirc\Box(p \rightarrow \$))$ to reward every state in which $p$ is true following the first issue of the command. To reward only the first occurrence $p$ after each command, we write $\Box(c \rightarrow \bigcirc(\neg p \, \mathsf{U} \, (p \wedge \$)))$. As for bounded variants for which we only reward goal achievement within $k$ steps of the command, we write for example $\Box(c \rightarrow \bigcirc_{\leq k}(p \rightarrow \$))$ to reward all such states in which $p$ holds.

It is also worth noting how to express simple behaviors involving past tense operators. To stipulate a reward if $p$ has always been true, we write $\$\,\mathsf{U}\,\neg p$. To say that we are rewarded if $p$ has been true since $q$ was, we write $\Box(q \rightarrow (\$ \, \mathsf{U} \, \neg p))$.

---
[2]If there is no $B$ such that $\models_B f$, which is the case for any $-free $f$ which is not a logical theorem, then $B_f$ is $\bigcap \emptyset$ – i.e. $S^*$. This limiting case is a little artificial, but since such formulae do not describe the attribution of rewards, it does no harm.

### 3.3 REWARD NORMALITY

$FLTL is so expressive that it is possible to write formulae which describe "unnatural" allocations of rewards. For instance, they may make rewards depend on future behaviors rather than on the past, or they may leave open a choice as to which of several behaviors is to be rewarded.[3] An example of the former is $\bigcirc p \rightarrow \$$, which stipulates a reward *now* if $p$ is going to hold *next*. We call such formula *reward-unstable*. What a reward-stable $f$ amounts to is that whether a particular prefix needs to be rewarded in order to make $f$ true does not depend on the future of the sequence. An example of the latter is $\Box(p \rightarrow \$) \vee \Box(\neg p \rightarrow \$)$ which says we should *either* reward all achievements of the goal $p$ *or* reward achievements of $\neg p$ but does not determine which. We call such formula *reward-indeterminate*. What a reward-determinate $f$ amounts to is that the set of behaviors modelling $f$, i.e. $\{B \mid \models_B f\}$, has a unique minimum. If it does not, $B_f$ is insufficient (too small) to make $f$ true.

In (Thiébaux et al., 2002), we show that formulae that are both reward-stable and reward-determinate – we call them *reward-normal* – are precisely those that capture the notion of "no funny business". This is this intuition that we ask the reader to note, as it will be needed in the rest of the paper. Just for reference then, we define:

**Definition 3** $f$ *is* reward-normal *iff for every* $\Gamma \in S^\omega$ *and every* $B \subseteq S^*$ $\Gamma \models_B f$ *iff* $B_f \cap \mathrm{pre}(\Gamma) \subseteq B$.

While reward-abnormal formulae may be interesting, for present purposes we restrict attention to reward-normal ones. Naturally, all formulae in Section 3.2 are normal.

### 3.4 $FLTL FORMULA PROGRESSION

Having defined a language to represent behaviors to be rewarded, we now turn to the problem of computing, given a reward formula, a minimum allocation of rewards to states actually encountered in an execution sequence, in such a way as to satisfy the formula. Because we ultimately wish to use anytime solution methods which generate state sequences incrementally via forward search, this computation is best done on the fly, while the sequence is being generated. We therefore devise an incremental algorithm inspired from a model-checking technique normally used to check whether a state sequence is a model of an FLTL formula (Bacchus and Kabanza, 1998). This technique is known as formula *progression* because it 'progresses' or 'pushes' the formula through the sequence.

In essence, our progression algorithm computes the modelling relation $\models_B$ given in Section 3.1, but unlike the def-

---
[3]These difficulties are inherent in the use of linear-time formalisms in contexts where the principle of directionality must be enforced. They are shared for instance by formalisms developed for reasoning about actions such as the Event Calculus and LTL action theories, see e.g. (Calvanese et al., 2002).

inition of $\models_B$, it is designed to be useful when states in the sequence become available one at a time, in that it defers the evaluation of the part of the formula that refers to the future to the point where the next state becomes available. Let $\Gamma_i$ be a state, say the last state of the sequence prefix $\Gamma(i)$ that has been generated so far, and let $b$ be a boolean true iff $\Gamma(i)$ is in the behavior $B$ to be rewarded. The progression of the $FLTL formula $f$ through $\Gamma_i$ given $b$, written $\text{Prog}(b, \Gamma_i, f)$, is a new *formula* satisfying the following property. Where $b \Leftrightarrow (\Gamma(i) \in B)$, we have:

**Property 1** $(\Gamma, i) \models_B f$ iff $(\Gamma, i+1) \models_B \text{Prog}(b, \Gamma_i, f)$

That is, given that $b$ tells us whether or not to reward $\Gamma(i)$, $f$ holds at $\Gamma_i$ iff the new formula $\text{Prog}(b, \Gamma_i, f)$ holds at the next (yet unavailable) state $\Gamma_{i+1}$ in the sequence. The function Prog is defined below:

**Algorithm 1** $FLTL Progression

| | |
|---|---|
| Prog(true, $s$, \$) | = $\top$ |
| Prog(false, $s$, \$) | = $\bot$ |
| Prog($b$, $s$, $\top$) | = $\top$ |
| Prog($b$, $s$, $\bot$) | = $\bot$ |
| Prog($b$, $s$, $p$) | = $\top$ iff $p \in s$ and $\bot$ otherwise |
| Prog($b$, $s$, $\neg p$) | = $\top$ iff $p \notin s$ and $\bot$ otherwise |
| Prog($b$, $s$, $f_1 \wedge f_2$) | = Prog($b$, $s$, $f_1$) $\wedge$ Prog($b$, $s$, $f_2$) |
| Prog($b$, $s$, $f_1 \vee f_2$) | = Prog($b$, $s$, $f_1$) $\vee$ Prog($b$, $s$, $f_2$) |
| Prog($b$, $s$, $\bigcirc f$) | = $f$ |
| Prog($b$, $s$, $f_1 \cup f_2$) | = Prog($b$, $s$, $f_2$) $\vee$ (Prog($b$, $s$, $f_1$) $\wedge$ $f_1 \cup f_2$) |
| Rew($s$, $f$) | = true iff Prog(false, $s$, $f$) = $\bot$ |
| \$Prog($s$, $f$) | = Prog(Rew($s$, $f$), $s$, $f$) |

This is to be matched with the definition of $\models_B$ in Section 3.1. Whenever $\models_B$ evaluates a subformula whose truth only depends on the current state, Prog does the same and return the formula $\top$ (resp. $\bot$) accordingly. Whenever $\models_B$ evaluates a subformula whose truth depends on future states, Prog defers the evaluation by returning a new subformula to be evaluated in the next state. Note that Prog is computable in linear time in the length of $f$, and that for \$-free formulae, it collapses to FLTL formula progression (Bacchus and Kabanza, 1998), regardless of the value of $b$.

Like $\models_B$, the function Prog assumes that $B$ is known, but of course we only have $f$ and one new state at a time of $\Gamma$, and what we really want to do is *compute* the appropriate $B$, namely that represented by $f$. So, similarly as in Section 3.1, we now turn to the second step, which is to use Prog to decide on the fly whether a newly generated sequence prefix $\Gamma(i)$ is in $B_f$ and so should be allocated a reward. This amounts to incrementally computing $B_f \cap \text{pre}(\Gamma)$, which provided $f$ is reward normal, is the minimal behavior $B$ such that $(\Gamma, 0) \models_B f$. We can do this as follows. According to Property 1,

$(\Gamma, 0) \models_B f$ iff $(\Gamma, 1) \models_B \text{Prog}(b_0, \Gamma_0, f_0)$, where $f_0 = f$ and $b_0$ stands for $\Gamma(0) \in B$. So B must be such that $(\Gamma, 1) \models_B \text{Prog}(b_0, \Gamma_0, f_0)$. To ensure minimality, we first assume that $\Gamma(0) \notin B$, i.e. $b_0$ is false, and compute Prog(false, $\Gamma_0, f_0$). If the result is $\bot$, then since no matter what $\Gamma_1$ turns out to be $(\Gamma, 1) \not\models_B \bot$, we know that the assumption about $b_0$ being false does not suffice to satisfy $f$. The only way to get $f$ to hold is to assign a reward to $\Gamma(0)$, so we take $\Gamma(0)$ to be in $B$, i.e. $b_0$ is true, and set the formula to be considered in the next state to $f_1 = \text{Prog}(\text{true}, \Gamma_0, f_0)$. If on the other hand the result is not $\bot$, then we need not reward $\Gamma(0)$ to make $f$ hold, so we take $\Gamma(0)$ not to be in $B$ and set $f_1 = \text{Prog}(\text{false}, \Gamma_0, f_0)$.

When $\Gamma_1$ becomes available, we can iterate this reasoning to compute the smallest value of $b_1$ such that $(\Gamma, 1) \models_B f_1$ and that of the corresponding $f_2 = \text{Prog}(b_1, \Gamma_1, f_1)$. And so on: progression through a sequence of states defines a sequence of booleans $\langle b_0, b_1, \ldots \rangle$ and a sequence of formulae $\langle f_0, f_1, \ldots \rangle$. When $\Gamma_i$ becomes available, we can compute the smallest value of $b_i$ such that $(\Gamma, i) \models_B f_i$ and the corresponding $f_{i+1}$. The value of $b_i$ represents $\Gamma(i) \in B_f$ and tells us whether we should allocate a reward at that stage, while $f_{i+1}$ is the new formula with which to iterate the process. In Algorithm 1, the function Rew takes $\Gamma_i$ and $f_i$ as parameter, and returns $b_i$ by computing the value of Prog(false, $\Gamma_i, f_i$). The function $Prog takes $\Gamma_i$ and $f_i$ as parameters and returns $f_{i+1}$ by calling Prog($b_i, \Gamma_i, f_i$) with the value of $b_i$ is given by Rew($\Gamma_i, f_i$).

The following theorem states that under weak assumptions, rewards are correctly allocated by progression:

**Theorem 1** *Let $f$ be reward-normal, and let $\langle f_0, f_1, \ldots \rangle$ be the result of progressing it through the successive states of a sequence $\Gamma$. Then, provided no $f_i$ is $\bot$, for all $i$ Rew($\Gamma_i, f_i$) iff $\Gamma(i) \in B_f$.*

The premise of the theorem is that $f$ does not eventually progress to $\bot$. Indeed if $f_i = \bot$ for some $i$, it means that even rewarding $\Gamma(i)$ does not suffice to make $f$ true, so something must have gone wrong: at some earlier stage, the boolean $b$ was made false where it should have been made true. The usual explanation is that the original $f$ was not reward-normal. For instance $\bigcirc p \to \$$, which is reward unstable, progresses to $\bot$ in the next state if p is true there: regardless of $\Gamma_0$, $f_0 = \bigcirc p \to \$ = \bigcirc \neg p \vee \$$, $b_0 = $ false, and $f_1 = \neg p$, so if $p \in \Gamma_1$ then $f_2 = \bot$. However, other (admittedly bizarre) possibilities exist: for example, although $\bigcirc p \to \$$ is reward-unstable, its substitution instance $\bigcirc \bigcirc \top \to \$$, which also progresses to $\bot$ in a few steps, is logically equivalent to \$ and is reward-normal.

If the progression method is to deliver the correct minimal behavior in all cases (even in all reward-normal cases) it has to backtrack on the choice of values for the $b_i$s. In the interest of efficiency, we choose not to allow backtracking. Instead, our algorithm raises an exception whenever a reward formula progresses to $\perp$, and informs the user of the sequence which caused the problem. The onus is thus placed on the domain modeller to select sensible reward formulae so as avoid possible progression to $\perp$. It should be noted that in the worst case, detecting reward-normality cannot be easier than the decision problem for $FLTL so it is not to be expected that there will be a simple syntactic criterion for reward-normality. In practice, however, commonsense precautions such as avoiding making rewards depend explicitly on future tense expressions suffice to keep things normal in all routine cases.

### 3.5 REWARD FUNCTIONS

With the language defined so far, we are able to compactly represent behaviors. The extension to a non-Markovian reward function is straightforward. We represent such a function by a set $\phi \subseteq \$FLTL \times \mathbb{R}$ of formulae associated with real valued rewards. We call $\phi$ a *reward function specification*. Where formula $f$ is associated with reward $r$ in $\phi$, we write '$(f : r) \in \phi$'. The rewards are assumed to be independent and additive, so that the reward function $R_\phi$ represented by $\phi$ is given by:

**Definition 4** $R_\phi(\Gamma(i)) = \sum_{(f:r) \in \phi} \{r \mid \Gamma(i) \in B_f\}$

E.g, if $\phi$ is $\{\neg p \cup p \wedge \$ : 5.2, \Box(q \rightarrow \Box\$) : 7.3\}$, we get a reward of 5.2 the first time that $p$ holds, a reward of 7.3 from the first time that $q$ holds onwards, a reward of 12.5 when both conditions are met, and 0 in otherwise.

Again, we can progress a reward function specification $\phi$ to compute the reward at all stages i of $\Gamma$. As before, progression defines a sequence $\langle \phi_0, \phi_1, \ldots \rangle$ of reward function specifications, with $\phi_{i+1} = \text{SProg}(\Gamma_i, \phi_i)$, where SProg is the function that applies Prog to all formulae in a reward function specification:

$$\text{SProg}(s, \phi) = \{(\text{Prog}(s, f) : r) \mid (f : r) \in \phi\}$$

Then, the total reward received at stage $i$ is simply the sum of the real-valued rewards granted by the progression function to the behaviors represented by the formulae in $\phi_i$:

$$\sum_{(f:r) \in \phi_i} \{r \mid \text{Rew}(\Gamma_i, f)\}$$

By proceeding that way, we get the expected analog of Theorem 1, which states progression correctly computes non-Markovian reward functions:

**Theorem 2** *Let $\phi$ be a reward-normal[4] reward function specification, and let $\langle \phi_0, \phi_1 \ldots \rangle$ be the result of progressing it through the successive states of a sequence $\Gamma$. Then, provided $(\perp : r) \notin \phi_i$ for any $i$, then*
$$\sum_{(f:r) \in \phi_i} \{r \mid \text{Rew}(\Gamma_i, f)\} = R_\phi(\Gamma(i)).$$

---
[4] We extend the definition of reward-normality to reward specification functions the obvious way, by requiring that all reward formulae involved be reward normal.

## 4 SOLVING NMRDPs

### 4.1 TRANSLATION INTO XMDP

We now exploit the compact representation of a non-Markovian reward function as a reward function specification to translate an NMRDP into an equivalent XMDP amenable to state-based anytime solution methods. Recall from Section 2.3 that each e-state in the XMDP is labeled by a state of the NMRDP and by history information sufficient to determine the immediate reward. In the case of a compact representation as a reward function specification $\phi_0$, this additional information can be summarized by the progression of $\phi_0$ through the sequence of states passed through. So an e-state will be of the form $\langle s, \phi \rangle$, where $s \in S$ is a state, and $\phi \subseteq \$FLTL \times \mathbb{R}$ is a reward function specification (obtained by progression). Two e-states $\langle s, \phi \rangle$ and $\langle t, \psi \rangle$ are equal if $s = t$, the immediate rewards are the same, and the results of progressing $\phi$ and $\psi$ through $s$ are semantically equivalent.

**Definition 5** *Let $D = \langle S, s_0, A, \Pr, R \rangle$ be an NMRDP, and $\phi_0$ be a reward function specification representing $R$ (i.e., $R_{\phi_0} = R$, see Definition 4). We translate $D$ into the XMDP $D' = \langle S', s'_0, A', \Pr', R' \rangle$ defined as follows:*

1. $S' \subseteq S \times 2^{\$FLTL \times \mathbb{R}}$

2. $s'_0 = \langle s_0, \phi_0 \rangle$

3. $A'(\langle s, \phi \rangle) = A(s)$

4. *If $a \in A'(\langle s, \phi \rangle)$, then $\Pr'(\langle s, \phi \rangle, a, \langle s', \phi' \rangle) =$*
$$\begin{cases} \Pr(s, a, s') & \text{if } \phi' = \text{SProg}(s, \phi) \\ 0 & \text{otherwise} \end{cases}$$

   *If $a \notin A'(\langle s, \phi \rangle)$, then $\Pr'(\langle s, \phi \rangle, a, \bullet)$ is undefined.*

5. $R'(\langle s, \phi \rangle) = \sum_{(f:r) \in \phi} \{r \mid \text{Rew}(s, f)\}$

Item 1 says that the e-states are labeled by a state and a reward function specification. Item 2 says that the initial e-state is labeled with the initial state and with the original reward function specification. Item 3 says that an action is applicable in an e-state if it is applicable in the state labeling it. Item 4 explains how successor e-states are and their probabilities are computed. Given an action $a$ applicable in an e-state $\langle s, \phi \rangle$, each successor e-state will be labeled by a successor state $s'$ of $s$ via $a$ in the NMRDP and by the progression of $\phi$ through $s$. The probability of that e-state is $\Pr(s, a, s')$ as in the NMRDP. Note that the cost of computing $\Pr'$ is linear in that of computing $\Pr$ and in the sum of the lengths of the formulae in $\phi$. Item 5 has been motivated before (see Section 3.5).

It is easy to show that this translation leads to an equivalent XMDP, as defined in Definition 1. Obviously, the function $\tau$ required for Definition1 is given by $\tau(\langle s, \phi \rangle) = s$, and then the proof is a matter of checking conditions.

## 4.2 BLIND MINIMALITY

The size of the XMDP obtained, i.e. the number of e-states it contains is a key issue for us, as it has to be amenable to state-based solution methods. Ideally, we would like the XMDP to be of minimal size. However, we do not know of a method building the *minimal* equivalent XMDP incrementally, adding parts as required by the solution method. And since in the worst case even the minimal XMDP can be larger than the NMRDP by a factor exponential in the length of the reward formulae (Bacchus et al., 1996), constructing it entirely would nullify the interest of anytime solution methods.

However, as we now explain, Definition 5 leads to an equivalent XMDP exhibiting a relaxed notion of minimality, and which is amenable to incremental construction. By inspection, we may observe that wherever an e-state $\langle s, \phi \rangle$ has a successor $\langle s', \phi' \rangle$ via action $a$, this means that in order to succeed in rewarding the behaviors described in $\phi$ by means of execution sequences that start by going from $s$ to $s'$ via $a$, it is necessary that the future starting with $s'$ succeeds in rewarding the behaviors described in $\phi'$. If $\langle s, \phi \rangle$ is in the minimal equivalent XMDP, and if there really are such execution sequences succeeding in rewarding the behaviors described in $\phi$, then $\langle s', \phi' \rangle$ must also be in the minimal XMDP. That is, construction by progression can only introduce e-states which are *a priori* needed. Note that an e-state that is *a priori* needed may not *really* be needed: there may in fact be no execution sequence using the available actions that exhibits a given behavior. For instance, consider the response formula $\Box(p \rightarrow \bigcirc^k q \rightarrow \bigcirc^k \$)$, i.e., every time command $p$ is issued, we will be rewarded $k$ steps later provided $q$ is true then. Obviously, whether $p$ is true at some stage affects the way future states should be rewarded. However, if $k$ steps from there a state satisfying $q$ can never be reached, then *a posteriori* $p$ is irrelevant, and there was no need to label e-states differently according to whether $p$ was true or not. To detect such cases, we would have to look perhaps quite deep into feasible futures. Hence the relaxed notion which we call *blind minimality* does not always coincide with absolute minimality.

We now formalise the difference between true and blind minimality. To simplify notation (avoiding functions like the $\tau$ of Definition 1), we represent each e-state as a pair $\langle s, r \rangle$ where $s \in S$ and $r$ is a function from $S^*$ to $\mathbb{R}$ intuitively assigning rewards to sequences in the NMRDP starting from $s$. A given $s$ may be paired with several functions $r$ corresponding to relevantly different histories of $s$. The XMDP is minimal if every such $r$ is *needed* to distinguish between reward patterns in the *feasible* futures of $s$:

**Theorem 3** *Let $S'$ be the set of e-states in a minimal equivalent XMDP $D'$ for $D = \langle S, s_0, A, \Pr, R \rangle$. Then for each e-state $\langle s, r \rangle \in S'$ there exists a prefix $\Gamma(i) \in \widetilde{D}(s_0)$ such that $\Gamma_i = s$ and for all $\Delta \in S^*$:*

$$r\langle\Delta\rangle = \begin{cases} R(\Gamma(i-1); \Delta) & \text{if } \Delta \in \widetilde{D}(s) \\ 0 & \text{otherwise} \end{cases}$$

Blind minimality is similar, except that, since there is no looking ahead, no distinction can be drawn between feasible trajectories and others in the future of $s$:

**Definition 6** *Let $S'$ be the set of e-states in an equivalent XMDP $D'$ for an NMRDP $D = \langle S, s_0, A, \Pr, R \rangle$. $D'$ is blind minimal iff for each e-state $\langle s, r \rangle \in S'$ there exists a prefix $\Gamma(i) \in \widetilde{D}(s_0)$ such that $\Gamma_i = s$ and for all $\Delta \in S^*$:*

$$r\langle\Delta\rangle = \begin{cases} R(\Gamma(i-1); \Delta) & \text{if } \Delta_0 = s \\ 0 & \text{otherwise} \end{cases}$$

**Theorem 4** *Let $D'$ be the translation of $D$ as in Definition 5. $D'$ is a blind minimal equivalent XMDP for $D$.*

## 4.3 EMBEDDED SOLUTION/CONSTRUCTION

Blind minimality is essentially the best achievable with anytime state-based solution methods which typically extend their envelope one step forward without looking deeper into the future. Our translation into a blind-minimal XMDP can be trivially embedded in any of these solution methods. This will result in an 'on-line construction' of the XMDP: the method will entirely drive the construction of those parts of the XMDP which it feels the need to explore, and leave the others implicit. If time is short, a suboptimal or even incomplete policy may be returned, but only a fraction of the state and expanded state spaces will be constructed. Note that the solution method should raise an exception as soon as one of the reward formulae progresses to $\bot$, i.e., as soon as an expanded state $\langle s, \phi \rangle$ is built such that $(\bot : r) \in \phi$, since this acts as a detector of unsuitable reward function specifications.

To the extent enabled by blind minimality, our approach allows for a dynamic analysis of the reward formulae, much as in (Bacchus et al., 1997). Indeed, only the execution sequences realisable under a particular policy actually explored by the solution method contribute to the analysis of rewards for that policy. Specifically, the reward formulae generated by progression for a given policy are determined by the prefixes of the execution sequences realisable under this policy. This dynamic analysis is particularly useful, since relevance of reward formulae to particular policies (e.g. the optimal policy) cannot be detected a priori.

The forward-chaining planner TLPlan (Bacchus and Kabanza, 2000) introduced the idea of using FLTL to specify domain-specific *search control knowledge* and formula progression to prune unpromising sequential plans (plans violating this knowledge) from deterministic search spaces. This has been shown to provide enormous time gains, leading TLPlan to win the 2002 planning competition hand-tailored track. Because our approach is based on progression, it provides an elegant way to exploit search control knowledge, yet in the context of decision-theoretic planning. Here this results in a dramatic reduction of the fraction of the XMDP to be constructed and explored, and therefore in substantially better policies by the deadline.

We achieve this as follows. We specify, via a $-free formula $c_0$, properties which we know must be verified by paths feasible under *promising* policies. Then we simply progress $c_0$ alongside the reward function specification, making e-states triples $\langle s, \phi, c \rangle$ where $c$ is a $-free formula obtained by progression. To prevent the solution method to apply an action that leads to the control knowledge being violated, the action applicability condition (item 3 in Definition 5) becomes: $a \in A'(\langle s,\phi,c\rangle)$ iff $a \in A(s)$ and $c \neq \bot$ (the other changes are straightforward). For instance, the effect of the control knowledge formula $\Box(p \to \bigcirc q)$ is to remove from consideration any feasible path in which $p$ is not followed by $q$. This is detected as soon as violation occurs, when the formula progresses to $\bot$. Although this paper focuses on non-Markovian rewards rather than dynamics, it should be noted that $-free formulae can also be used to express non-Markovian constraints on the system's dynamics, which can be incorporated in our approach exactly as we do for the control knowledge.

## 5　RELATED AND FUTURE WORK

It is evident that our thinking about solving NMRDPs and the use of temporal logic to represent them draws on (Bacchus et al., 1996). Both this paper and (Bacchus et al., 1997) advocate the use of PLTL over a finite past to specify non-Markovian rewards. In the PLTL style of specification, we describe the past conditions under which we get rewarded now, while with $FLTL we describe the conditions on the present and future under which future states will be rewarded. While the behaviors and rewards may be the same in each scheme, the naturalness of thinking in one style or the other depends on the case. Letting the kids have a strawberry dessert because they have been good all day fits naturally into a past-oriented account of rewards, whereas promising that they may watch a movie if they tidy their room (indeed, making sense of the whole notion of promising) goes more naturally with FLTL. One advantage of the PLTL formulation is that it trivially enforces the principle that present rewards do not depend on future states. In $FLTL, this responsibility is placed on the domain modeller. On the other hand, the greater expressive power of $FLTL opens the possibility of considering a richer class of decision processes, e.g. with uncertainty as to which rewards are received (the dessert or the movie) and when (some time next week, before it rains). This is a topic for future work. At any rate, as we now explain, $FLTL is better suited than PLTL to solving NMRDPs using anytime state-based solution methods.

(Bacchus et al., 1996) proposes a method whereby an e-state is labeled by a set of subformulae of the PLTL reward formulae. For the labeling, two extreme cases are considered: one very simple and the other elaborate. In the simple case, an e-state is labeled by the set of all subformulae which are true at it. The computation of such simple labels can be done forward starting from the initial state, and so could be embedded in an anytime solution method. However, because the structure of the original reward formulae is lost when considering subformulae individually, fine distinctions between histories are drawn which are totally irrelevant to the reward function. Consequently, the expanded state space easily becomes exponentially bigger than the blind-minimal one. This is problematic with the solution methods we consider, because size severely affects their performance in solution quality.

In the elaborate case, a pre-processing phase uses PLTL formula regression to find sets of subformulae as potential labels for possible predecessor states, so that the subsequent generation phase builds an XMDP representing all and only the histories which make a difference to the way *actually feasible* execution sequences should be rewarded. The XMDP produced is minimal, and so in the best case exponentially smaller than the blind-minimal one. However, the prohibitive cost of the pre-processing phase makes it unusable for anytime solution methods (it requires exponential space and a number of iterations through the state space exponential in the size of the reward formulae). We do not consider that any method based on PLTL and regression will achieve a meaningful relaxed notion of minimality without a costly pre-processing phase. Our main contribution is an approach based on FLTL and progression which does precisely that, letting the solution method resolve the tradeoff between quality and cost in a principled way intermediate between the two extreme suggestions above.

The structured representation and solution methods targeted by (Bacchus et al., 1997) differ from the anytime state-based solution methods our framework primarily aims at, in particular in that they do not require explicit state enumeration at all (Boutilier et al., 2000; Hoey et al., 1999). Accordingly, the translation into XMDP given in (Bacchus et al., 1997) keeps the state and expanded state space implicit, and amounts to adding temporal variables to the problems together with the decision-tree describing their dynamics. It is very efficient but rather crude: the encoded history features do not even vary from one state to the next, which strongly compromises the minimality of the XMDP.[5] However, non-minimality is not as problematic as with the state-based approaches, since structured solution methods do not enumerate states and are able to dynamically ignore some of the variables that become irrelevant at some point of policy construction.

---

[5](Chomicki, 1995) uses a similar approach to extend a database with auxiliary relations containing additional information sufficient to check temporal integrity constraints. As there is only ever one sequence of databases, what matters is more the size of these relations than avoiding making redundant distinctions.

In another sense, too, our work represents a middle way, combining the advantages conferred by state-based and structured approaches, e.g. by (Bacchus et al., 1996) on one side, and (Bacchus et al., 1997) on the other. From the former we inherit a meaningful notion of minimality. As with the latter, approximate solution methods can be used and can perform a restricted dynamic analysis of the reward formulae. In virtue of the size of the XMDP produced, the translation proposed in (Bacchus et al., 1997) is clearly unsuitable to anytime state-based methods. The question of the appropriateness of our translation to structured solution methods, however, cannot be settled as clearly. On the one hand, our approach does not preclude the exploitation of a structured representation of *system's states*,[6] and formula progression enables even state-based methods to exploit *some* of the structure in '$FLTL space'. On the other hand, the gap between blind and true minimality indicates that progression alone is insufficient to always *fully* exploit that latter structure (reachability is not exploited). With our translation, even structured solution methods will not remedy this. There is a hope that (Bacchus et al., 1997) is able to take advantage of the full structure of the reward function, but also a possibility that it will fail to exploit even as much structure as our approach would, as efficiently.

The most important item for future work is an empirical comparison of the three approaches in view to answering this question and identifying the domain features favoring one over the other. Ours has been fully implemented as an extension (rewards, probabilities) of TLPlan's planning language (Bacchus and Kabanza, 2000), which, like TLPlan's, includes functions and bounded quantification. To allow for comparisons to be reported in a longer version of this paper, we are in the process of implementing the other two approaches, no implementation of either of them being reported in the literature. Another exciting future work area is the investigation of temporal logic formalisms for specifying heuristics for NMRDPs or more generally for search problems with temporally extended goals, as good heuristics are important to some of the solution methods we are targeting. Related to this is the problem of extending, to temporally extended goals, the GOALP predicate (Bacchus and Kabanza, 2000) which is the key to the specification of reusable control knowledge in the case of reachability goals. Finally, we should investigate the precise relationship between our line of work and recent work on planning for weak temporally extended goals in non-deterministic domains, such as attempted reachability and maintenance goals (Pistore and Traverso, 2002).

**Acknowledgements** Many thanks to Fahiem Bacchus, Rajeev Goré, Charles Gretton, David Price, Leonore Zuck, and reviewers for useful discussions and comments.

---

[6]Symbolic implementations of the solution methods we consider, e.g. (Feng and Hansen, 2002), as well as formula progression in the context of symbolic state representations (Pistore and Traverso, 2001) could be investigated for that purpose.